\documentclass[letterpaper]{article}
\usepackage{aaai}
\usepackage{times}
\usepackage{helvet}
\usepackage{courier}
\usepackage{epic}
\usepackage{ecltree}
\usepackage{graphicx}
\usepackage{latexsym}
\begin{document}

\newcommand{\set}{\mathcal}
\newcommand{\myset}[1]{\ensuremath{\mathcal #1}}

\renewcommand{\theenumii}{\alph{enumii}}
\renewcommand{\theenumiii}{\roman{enumiii}}
\newcommand{\figref}[1]{Figure \ref{#1}}
\newcommand{\tref}[1]{Table \ref{#1}}
\newcommand{\And}{\wedge}
\newcommand{\myldots}{.}

\newtheorem{mydefinition}{Definition}
\newtheorem{mytheorem}{Theorem}
\newtheorem{mytheorem1}{Theorem}
\newcommand{\myproof}{\noindent {\bf Proof:\ \ }}
\newcommand{\myqed}{\mbox{$\Box$}}

\newcommand{\mymod}{\mbox{\rm mod}}
\newcommand{\range}{\mbox{\sc Range}}
\newcommand{\roots}{\mbox{\sc Roots}}
\newcommand{\myiff}{\mbox{\rm iff}}
\newcommand{\alldifferent}{\mbox{\sc AllDifferent}}
\newcommand{\permutation}{\mbox{\sc Permutation}}
\newcommand{\disjoint}{\mbox{\sc Disjoint}}
\newcommand{\cardpath}{\mbox{\sc CardPath}}
\newcommand{\CARDPATH}{\mbox{\sc CardPath}}
\newcommand{\common}{\mbox{\sc Common}}
\newcommand{\uses}{\mbox{\sc Uses}}
\newcommand{\lex}{\mbox{\sc Lex}}
\newcommand{\usedby}{\mbox{\sc UsedBy}}
\newcommand{\nvalue}{\mbox{\sc NValue}}
\newcommand{\slide}{\mbox{\sc Slide}}
\newcommand{\SLIDE}{\mbox{\sc Slide}}
\newcommand{\circularslide}{\mbox{\sc Slide}_{\rm O}}
\newcommand{\among}{\mbox{\sc Among}}
\newcommand{\mysum}{\mbox{\sc MySum}}
\newcommand{\amongseq}{\mbox{\sc AmongSeq}}
\newcommand{\atmost}{\mbox{\sc AtMost}}
\newcommand{\atleast}{\mbox{\sc AtLeast}}
\newcommand{\element}{\mbox{\sc Element}}
\newcommand{\gcc}{\mbox{\sc Gcc}}
\newcommand{\gsc}{\mbox{\sc Gsc}}
\newcommand{\contiguity}{\mbox{\sc Contiguity}}
\newcommand{\PRECEDENCE}{\mbox{\sc Precedence}}
\newcommand{\precedence}{\mbox{\sc Precedence}}
\newcommand{\assignnvalues}{\mbox{\sc Assign\&NValues}}
\newcommand{\linksettobooleans}{\mbox{\sc LinkSet2Booleans}}
\newcommand{\domain}{\mbox{\sc Domain}}
\newcommand{\symalldiff}{\mbox{\sc SymAllDiff}}

\newcommand{\slidingsum}{\mbox{\sc SlidingSum}}
\newcommand{\MaxIndex}{\mbox{\sc MaxIndex}}
\newcommand{\REGULAR}{\mbox{\sc Regular}}
\newcommand{\regular}{\mbox{\sc Regular}}
\newcommand{\cfg}{\mbox{\sc Grammar}}
\newcommand{\CFG}{\mbox{\sc Grammar}}
\newcommand{\GRAMMAR}{\mbox{\sc Grammar}}
\newcommand{\valsym}{\mbox{\sc ValSym}}

\newcommand{\DFA}{\mbox{\sc DFA}}
\newcommand{\STRETCH}{\mbox{\sc STretch}}
\newcommand{\SLIDEOR}{\mbox{\sc SlideOr}}
\newcommand{\NAE}{\mbox{\sc NotAllEqual}}

\newcommand{\todo}[1]{{\tt (... #1 ...)}}

\title{Breaking Value Symmetry\thanks{The 
author is funded by
the Australian Government's Department of Broadband, Communications and the Digital Economy
and the 
Australian Research Council.}
}

\author{Toby Walsh\\
NICTA and UNSW\\
Sydney, Australia\\
toby.walsh@nicta.com.au}
%\Date{}
%\setlength\titlebox{2in}

%\date{1st January 2008}

\maketitle
\begin{abstract}
Symmetry is an important factor in solving many
constraint satisfaction problems. 
One common type of symmetry is when we have symmetric values. 
In a recent series of papers, 
we have studied methods to break value symmetries 
\cite{wcp06,wcp07}.
Our results identify computational
limits on eliminating value symmetry.
For instance, we prove that pruning all symmetric values 
is NP-hard in general. Nevertheless, 
experiments show that much value symmetry can be broken
in practice. These results may
be useful to researchers in planning, scheduling
and other areas as value symmetry occurs in many
different domains.
\end{abstract}

\section{Introduction}

Many search problems contain symmetries. Symmetry
occurs naturally in many problems
(e.g. if we have identical machines to schedule,
identical jobs to process, or equivalently skilled personnel
to roster).
Symmetry can also be introduced when we model a problem
(e.g. if we name the elements in a set, we
introduce the possibility of permuting their order). 
Unfortunately, symmetries increases the size of the
search space. If we do not eliminate symmetries, 
we will waste much time visiting
symmetric solutions, as well as
those parts of the search tree which
are symmetric to already visited states.

One common type of symmetry is when values are
symmetric. For example, if we are assigning
colors (values) to nodes (variables) in a graph coloring
problem, we can uniformly swap the names of the colors
throughout a coloring. 
As a second example if we are assigning nurses (values)
to shifts (variables) in a rostering problems, and two
nurses have the same skills, we may be able to interchange
them uniformly throughout the schedule. 
In a recent series of papers \cite{wcp06,wcp07}, 
we have studied methods to eliminate such value
symmetries. A clear picture
of breaking value symmetry is emerging from these
studies. These results may be useful to researchers
in other areas of AI as value symmetry is a common
feature of many domains. 

\section{An example}

To illustrate the ideas, we consider a simple
problem from musical composition. 
The all interval series problem (prob007 in CSPLib.org)
asks for a permutation 
of the numbers 0 to $n-1$
so that neighbouring differences
form a permutation of 1 to $n-1$. 
For $n=12$, the problem corresponds to 
arranging the half-notes of a scale
so that all musical intervals (minor second to
major seventh) are covered. 
This is a simple example of a graceful graph
problem in which the graph is a path.

We can model
this as a constraint satisfaction problem
in $n$ variables with
$X_i=j$ iff the $i$th number in the series is $j$. 
One solution for $n=11$ is:
\begin{eqnarray}
X_1, X_2, \ldots, X_{11} & = & 
3, 7, 4, 6, 5, 0, 10, 1, 9, 2, 8 
\end{eqnarray}
%The differences form the series: $ 4, 3, 2, 1, 5, 10, 9, 8, 7, 6 $.

The all interval series problem has a number of different symmetries.
First, we can reverse any solution and generate a new
(but symmetric) solution:
\begin{eqnarray}
X_1, X_2,   \ldots, X_{11} & = & 
 8, 2, 9, 1, 10, 0, 5, 6, 4, 7, 3
\end{eqnarray}
Second, the all interval series problem has a value symmetry 
as we can invert values.
If we subtract all values in (1) from $10$, we
generate a second (but symmetric) solution:
\begin{eqnarray}
X_1,  X_2,  \ldots, X_{11} & = & 
 7, 3, 6, 4, 5, 10, 0, 9, 1, 8, 2
\end{eqnarray}
Third, we can do both and generate a third (but symmetric)
solution:
\begin{eqnarray}
X_1,  X_2,  \ldots, X_{11} & = & 
 2, 8, 1, 9, 0, 10, 5, 4, 6, 3, 7
\end{eqnarray}

To eliminate such symmetric solutions from the
search space, we can post additional constraints
which eliminate all but one solution in each
symmetry class. 
To eliminate the reversal of a solution,
we can simply post the constraint:
\begin{eqnarray}
& X_1 < X_{11}&
\end{eqnarray}
This eliminates solution (2) as it is a reversal of (1). 

To eliminate the value symmetry which subtracts all values
from $10$, we can post:
\begin{eqnarray}
X_1 \leq 10-X_1, &  X_1=10-X_1 \Rightarrow X_2 < 10-X_2 &
\end{eqnarray}
This is equivalent to:
\begin{eqnarray*}
X_1 \leq 5, & & X_1=5 \Rightarrow X_2 < 5
\end{eqnarray*}
This eliminates solutions (2) and (3). 
Finally, eliminating the third symmetry
where we both reverse the solution and subtract it
from $10$ is more difficult. We can, for instance, post:
\begin{eqnarray}
& X_1 \leq 10-X_{11}, & \nonumber \\
& X_1=10-X_{11} \Rightarrow X_2 \leq 10-X_{10}, & \nonumber  \\
& X_1=10-X_{11} \ \& \ X_2=10-X_{10} \Rightarrow X_3 \leq 10-X_{9}, &  \nonumber \\
& \vdots  & 
\end{eqnarray}
Note that of the four symmetric solutions given
earlier, only (4)
with $X_1=2$, $X_2=8$ and $X_{11}=7$
satisfies all three sets of symmetry breaking constraints: (5), (6) and (7).
The other three solutions are eliminated.
This leaves the question 
of where symmetry breaking constraints like
(6) and (7) come from in general. Our work 
helps answer this question.

\section{Formal background}

A constraint satisfaction problem consists of a set of variables,
each with a domain of values, and a set of constraints
specifying allowed combinations of values for given subsets of
variables. Variables take one value from a given
finite set.
A solution is an assignment of values to variables
satisfying the constraints.
%Constraint solvers typically prune
%values for variables which cannot be in any
%solution. A constraint is \emph{domain consistent}
%iff for each variable, every value in its domain 
%can be extended to an assignment that satisfies
%the constraint. 
%
Symmetry occurs in many constraint satisfaction
problems. A \emph{value symmetry} is a permutation of the values
that preserves solutions. More formally,
a value symmetry is a bijective mapping, $\sigma$ of the
values such that if $X_1=d_1, \ldots, X_n=d_n$ is a solution
then $X_1=\sigma(d_1), \ldots, X_n=\sigma(d_n)$ is also. 
For example, in the all interval series
problem, the value symmetry $\sigma$ maps the value $i$ onto
$n-1-i$. 
A \emph{variable symmetry}, on the other hand,
is a permutation of the variables
that preserves solutions. 
More formally,
a variable symmetry is a bijective mapping, $\theta$ of the
indices of variables such that if $X_1=d_1, \ldots, X_n=d_n$ is a solution
then $X_{\theta(1)}=d_1, \ldots, X_{\theta(n)}=d_n$ is also. 
For example, in the all interval series
problem, the variable symmetry for reversing
a series maps the index $i$ onto $n-i+1$. 
Symmetries are problematic as they increase 
the size of the search space. 
For instance, $m$ interchangeable
values increases the size of the search space
by a factor of $m!$.

\section{Lex-Leader method}

One simple and effective mechanism to deal with symmetry is to
add constraints which eliminate symmetric solutions \cite{puget:Sym}.
For variable symmetries, Crawford {\it et al.} 
give a simple method that eliminates all symmetric solutions
\cite{clgrkr96}. 
The basic idea is very simple. 
We pick an ordering on the variables, and
then post symmetry breaking constraints to
ensure that the final solution is lexicographically
less than any symmetric re-ordering of the variables. 
For example, with a reversal symmetry which maps $X_i$ to $X_{n-i+1}$, 
we post the lexicographical ordering constraint:
\begin{eqnarray*}
[ X_1, \ldots , X_n]  & \leq_{\rm lex} & 
[ X_n, \ldots , X_1] 
\end{eqnarray*}
This selects the ``lex leader'' assignment. Note
that if $X_i$ are all-different, as they are in the all interval
series problem, we can simplify this lex leader
constraint to give as in (5):
\begin{eqnarray*}
& X_1  <   X_n &
\end{eqnarray*}

The lex-leader method lies at the heart
of most static methods for breaking
variable symmetry. 
In their survey of symmetry breaking in constraint
programming in the recent Handbook of Constraint
Programming, Gent, Petrie and Puget observe that:
\begin{quote}
``\ldots {\em lex-leader remains of the highest importance, because
there a number of ways it is used to derive new symmetry breaking
methods \ldots [however] the lex-leader method is defined only for variable
symmetries \ldots a proper generalisation of lex-leader to deal with
value symmetries would be valuable, even if restricted to
some special cases} \ldots'' page 345 \cite{symchap06}
\end{quote}

This is the research challenge we have
been tackling in our recent work \cite{wcp06,wcp07}.

\section{Value symmetry}

In \cite{wcp06}, we propose a generalization of the 
lex-leader method that
works with any type of value symmetry. 
Suppose we have a set $\Sigma$ of value symmetries. 
We can eliminate
all symmetric solutions with
the constraint
$\valsym(\Sigma,[X_1, \ldots , X_n])$
which ensures 
for all
$\sigma \in \Sigma$:
\begin{eqnarray}
[ X_1, \ldots , X_n]  & \leq_{\rm lex} & 
[ \sigma(X_1), \ldots , \sigma(X_n)] 
\end{eqnarray}
Where $X_1$ to $X_n$ is any fixed ordering
on the variables and $\sigma(X_i)$ represents the action
of the symmetry $\sigma$ on the value assigned to $X_i$. 
For example, if $\sigma$ inverts values
by mapping $i$ onto $n-1-i$ as in the all interval
series, we can simplify (8) to give:
\begin{eqnarray*}
[ X_1, X_2, \ldots ]  & \leq_{\rm lex} & 
[ n-1-X_1, n-1-X_2, \ldots] 
\end{eqnarray*}
Expanding the definition of lex, and exploiting
the fact that $X_1 \neq X_2$, we get as in (6):
\begin{eqnarray*}
 X_1 \leq n-1-X_1,  & X_1=n-1-X_1 \Rightarrow X_2 < n-1-X_2
\end{eqnarray*}
We give a linear time propagator for
lex-leader constraints like (8), and show how
to extend the lex-leader method
to symmetries which act on both variables
and values simultaneously. % \cite{wcp06}. 

In theory, this generalization of the lex-leader method
solves the problem of value symmetries as it
eliminates all symmetric solutions. %, and helping to prune
%symmetric states. 
Unfortunately, several problems remain.
First, the set of 
symmetries $\Sigma$ can be exponentially large (for example,
there are $m!$ symmetries if we have $m$ interchangeable
values) requiring us to post a large number
of lex ordering constraints.
Second, decomposing $\valsym$ into
separate lex ordering constraints 
hinders propagation and may prevent
all symmetric values from being pruned. In \cite{wcp06},
we give a simple example where this decomposition
hinders propagation. 
Somewhat surprisingly, 
there is little hope to overcome
this second problem.
We prove in \cite{wcp07} that
pruning all symmetric values
is intractable in general
even with a small number of symmetries
(assuming $P \neq NP$).

\begin{mytheorem}
Pruning
all inconsistent values
for $\valsym(\Sigma,[X_1, \myldots , X_n])$ is 
NP-hard, even when $|\Sigma|$ is linearly bounded. 
\end{mytheorem}
\myproof
In \cite{wcp07}, we reduce 3-SAT to deciding
if a particular \valsym\ constraint has a
solution. 
\myqed

With value symmetry,
whilst pruning all symmetric values is NP-hard, 
method likes those in \cite{getree,pcp05} will 
eliminate all symmetric solutions in polynomial time. 
Despite the negative result given in Theorem 1,
value symmetry appears easier to break than variable symmetry
both in theory and in practice. 
For example, with variable symmetry, we 
prove in \cite{bhhwaaai2004}
that just eliminating all symmetric solutions is 
itself NP-hard.

\section{Interchangeable values}

So far, we have considered symmetries in general
and ignored any special properties of the symmetry group. 
A common type of value symmetry with special properties
that we can exploit is that due to interchangeable values.
We can break all such value symmetry
using value {\em precedence} \cite{llcp2004},
an idea which has been used in many contexts including
the least number heuristic \cite{lnh}. 
To ensure value precedence, we can use
the global constraint $\PRECEDENCE([X_1,\myldots ,X_n])$.
This 
holds iff %$\min \{ i \ | \ X_i=j \vee i=n+1\} < 
% \min \{ i \ | \ X_i=k \vee i=n+2\}$ for all $j<k$. 
%That is, 
the first time we use $j$ is before
the first time we use $k$ for all $j<k$. 
For example, consider an assignment like:
\begin{eqnarray}
X_1, X_2, X_3, \ldots, X_n & = & 1, 1, 2, 1, 3, \ldots, 2
\end{eqnarray}
This satisfies value precedence as 1 first occurs
before 2, 2 first occurs before 3, etc. 
Now consider the symmetric assignment in which we swap 
$2$ with $3$:
\begin{eqnarray}
X_1, X_2, X_3, \ldots, X_n & = & 1, 1, 3, 1, 2, \ldots, 3
\end{eqnarray}
This does not satisfy value precedence as 
$3$ first occurs before $2$. 
Posting a \PRECEDENCE\ constraint 
eliminates all symmetric solutions due to interchangeable values. 
In \cite{wecai2006}, we give a linear time propagator
for the \PRECEDENCE\ constraint.
In \cite{wcp07}, 
we argue that $\PRECEDENCE$ is 
in fact equivalent to
$\valsym$. 

Another way to ensure value precedence
is to channel into dual variables, $Z_j$ which record the
first index using each value $j$
\cite{pcp05}. 
This maps value symmetry into variable
symmetry on the $Z_j$. We can then break
this variable symmetry by posting simple
ordering constraints:
\begin{eqnarray}
& Z_1 < Z_2 < Z_3 < \ldots < Z_m & 
\end{eqnarray}
This ensures that the first occurrence of 1 is
before that of 2, that of 2 is before
3, etc.
For example, the assignment in (9) 
satisfies (11) as $Z_1=1$, $Z_2=3$ and $Z_3=5$.
However, the symmetric assignment in (10)
does not satisfy (11) as $Z_2=5$ but $Z_3=3$. 
Puget proves that we can, in fact, break
{\em any} value symmetry with a linear
number of ordering constraints on the $Z_j$.
Unfortunately, even with just two value symmetries,
Puget's method hinders propagation \cite{wcp07}. 
%even for the tractable case of interchangeable values. 
%Indeed, even with \emph{just} two value symmetries,
%mapping into variable symmetry can hinder propagation. 
This is supported by the experiments in \cite{wcp07}
where we see faster and more effective symmetry
breaking with the global \PRECEDENCE\ constraint. 
This is therefore a promising method to eliminate 
the symmetry due to interchangeable values. 

\section{Dynamic methods}

An alternative to static methods which
add constraints to eliminate 
symmetric solutions are dynamic methods
which modify
the search procedure to ignore 
symmetric branches. 
For example, with value symmetries,
the GE-tree method dynamically
eliminates all symmetric solutions
from a backtracking search procedure 
in $O(n^4 \log(n))$ time \cite{getree}. 
Dynamic methods have the advantage
that they do not conflict with the
branching heuristic. However, dynamic methods
do not prune the search space as much.
%may not prune
%as many symmetric values as 
%static methods. 
%
Suppose we are at a particular node
in the search tree explored by the GE-tree method. 
%Consider the current and all past variables seen so far. 
The GE-tree method essential
performs only forward checking, 
pruning symmetric assignments
from the domain of the {\em next} branching 
variable. 
Unlike static
methods, the GE-tree method 
does not prune {\em deeper} variables.
By comparison, static symmetry breaking
constraints can prune deeper
variables, resulting in
interactions between the problem constraints and additional
domain prunings. For this reason, 
static symmetry breaking methods can
solve certain problems exponentially quicker
than dynamic methods. 
\begin{mytheorem}
  There exists a model of the pigeonhole problem in $n$ variables and
  $n+1$ interchangeable values 
  such that, 
  given any variable and value ordering, the GE-tree
  method explores $O(2^n)$ branches, but which 
  static symmetry breaking 
  methods like value precedence solve in just $O(n^2)$ time.
\end{mytheorem}
\myproof
See \cite{wcp07}.
\myqed

This theoretical result supports the experimental results
in \cite{pcp05} showing that static methods for breaking
value symmetry outperform dynamic methods. Nevertheless,
dynamic methods have been found useful as they may
not conflict with the branching heuristic \cite{grcp07,vhmcpaior08}. 
An interesting direction for future work are
hybrid methods that combine the best features of
static and dynamic symmetry breaking.

\section{Related work}

Puget proved that symmetric solutions can be eliminated
by the addition of suitable constraints \cite{puget:Sym}.
Crawford {\it et al.}  presented
the first general method for constructing 
variable symmetry breaking constraints \cite{clgrkr96}.
To deal with large number of symmetries, 
Aloul {\it et al.} suggest breaking only
those symmetries which are generators \cite{armsdac2002}. 
%Aloul {\it et al.} also 
%improved the runtime of this method by reducing the size
%of a CNF encoding of such a symmetry breaking constraint
%from quadratic to linear \cite{faijcai03}. 
Puget and Walsh independently proposed propagators for
value symmetry breaking constraints \cite{paaai2006,wcp06}. 
To deal with the exponential 
number of such value symmetry breaking constraints,
Puget proposed a global propagator 
which does forward checking in polynomial time \cite{paaai2006}.

To eliminate symmetric solutions due to interchangeable values, 
Law and Lee 
defined value precedence for
finite domain and set variables 
and proposed a specialized propagator for 
a pair of interchangeable values
\cite{llcp2004}. 
Walsh extended this to a propagator for
any number of interchangeable values \cite{wecai2006}. 
Value precedence enforces
the so-called ``least number heuristic''
\cite{lnh}. 
Finally, an alternative way to break
value symmetry statically is to convert it into a variable
symmetry by channelling into a dual Boolean viewpoint 
in which $B_{ij}=1$ iff $X_i=j$, 
and using lexicographical ordering constraints on the
Boolean variables \cite{ffhkmpwcp2002,llconstraints06}. 
More recently, 
static symmetry breaking constraints have
been proposed to eliminate the symmetry due to both interchangeable 
variables and values \cite{fpsvcp06,llwycp07}.

A number of dynamic methods 
have been proposed to deal with value symmetry.
Van Hentenryck {\it et al.} gave a 
labelling schema for
eliminating all symmetric solutions due to interchangeable values
\cite{hafpijcai2003}. 
Inspired by this method, 
Roney-Dougal {\it et al.} 
gave a polynomial method to construct 
a GE-tree, a search tree without value symmetry \cite{getree}. 
Finally, Sellmann and van Hentenryck gave
a $O(nd^{3.5}+n^2d^2)$
dominance detection algorithm
for eliminating all symmetric solutions when both 
variables and values are interchangeable
\cite{sellmann2}.

\section{Conclusions}

Value symmetries can be broken either statically (by
adding constraints to prune symmetric solutions) 
or dynamically (by modifying the search procedure to 
avoid symmetric branches). We have shown that
both approaches have computational limitations. 
With static methods, we can eliminate
all symmetric solutions in polynomial time but
pruning all symmetric values
is NP-hard in general (or equivalently, we can
avoid visiting symmetric leaves of the search
tree in polynomial time but avoiding symmetric subtrees is
NP-hard).
With dynamic methods, we can take
exponential time on problems which static methods
solve without search. Nevertheless, 
experimental results in \cite{paaai2006,wecai2006,wcp06,wcp07}
and elsewhere show that value symmetry can be dealt with effectively
in practice. 

These results may be useful to researchers
in closely related areas like planning
and scheduling where value symmetries arise
and need to be eliminated. They may also
be useful in areas like model counting,
verification, and automated reasoning as 
value symmetry arises in many of 
these problems too. 
There are many open questions raised by this
research. For example,
are there symmetries 
where all symmetric values can be pruned tractably?  
How do we combine the best features of static and dynamic 
symmetry breaking?

\bibliographystyle{aaai}
%%\bibliographystyle{alpha}
%%\bibliography{/home/s5/tw/biblio/a-z,/home/s5/tw/biblio/pub}
\bibliography{/home/tw/biblio/a-z,/home/tw/biblio/pub,/home/tw/biblio/a-z2,/home/tw/biblio/pub2}
%\bibliography{/Users/twalsh/Documents/biblio/a-z,/Users/twalsh/Documents/biblio/pub}
%%\bibliography{/n/endjinn/u6/tw/biblio/a-z,/n/endjinn/u6/tw/biblio/pub}
%\bibliography{/usr/tw/biblio/a-z,/usr/tw/biblio/pub}
%\bibliography{references.bib}
%%\bibliography{/home/arp/disk1/tw/biblio/a-z,/home/arp/disk1/tw/biblio/pub}
%%\bibliography{/u6/tw/biblio/a-z,/u6/tw/biblio/pub}
%%\bibliography{/usr/local/users/tw/biblio/a-z,/usr/local/users/tw/biblio/pub}
%\bibliography{biblio}

%\end{thebibliography}

\end{document}